\documentclass[12pt]{amsart}

\usepackage[textheight=23.5cm, left=1.2cm, right=1.2cm]{geometry}

\usepackage[english]{babel}
\usepackage[T1]{fontenc}
\usepackage[utf8]{inputenc}

\theoremstyle{plain}
    \newtheorem{theorem}{Theorem}[section]
    \newtheorem{lemma}[theorem]{Lemma}

    \newtheorem{question}[theorem]{Question}

\theoremstyle{definition}

\usepackage[
    draft = false,
    unicode = true,
    colorlinks = true,
    allcolors = blue,
    hyperfootnotes = true,
    citecolor = red
]{hyperref}
\usepackage{amsfonts,amsmath,amssymb,amscd,amsthm,url}
\usepackage[inline]{enumitem}
\usepackage{graphicx,epsf,afterpage}
\usepackage{multicol}
\usepackage{array}
\usepackage{braket}
\usepackage{epigraph}
\usepackage{rotating}

\usepackage{xcolor}

\usepackage{ulem}

\usepackage{tikz}
\usetikzlibrary{positioning, calc, intersections, through, arrows, matrix, chains}
\usetikzlibrary{graphs}
\usetikzlibrary{graphs.standard}
\usetikzlibrary{patterns}
\usetikzlibrary{decorations.pathreplacing,calligraphy,backgrounds}

\newcommand\norm[1]{\ensuremath{\left\lVert#1\right\rVert}}
\newcommand\abs[1]{\ensuremath{\left\lvert#1\right\rvert}}

\newcommand{\R}{\ensuremath{\mathbb{R}}}

\newcommand{\Q}{\ensuremath{\mathbb{Q}}}

\renewcommand{\geq}{\geqslant}

\renewcommand{\leq}{\leqslant}
\renewcommand{\emptyset}{\varnothing}

\newcounter{mcnt}

\newcounter{wordcnt}

\usepackage[
backend=biber,
style=numeric-comp,
sorting=none
]{biblatex}
\usepackage{csquotes}

\tikzset{>=stealth',every on chain/.append style={join},
         every join/.style={->}}
\tikzstyle{labeled}=[execute at begin node=$\scriptstyle,
   execute at end node=$]

\DeclareMathOperator{\appx}{Approx}

\begin{document}

\title{Kolmogorov--Arnold stability}
\author{Sviatoslav V. Dzhenzher and Michael H. Freedman}

\thanks{\hspace{-5mm}
S.~V.~Dzhenzher: Moscow Institute of Physics and Technology, Dolgoprudny 141701, Russia. sdjenjer@yandex.ru
\\
M.~H.~Freedman: (Senior Research Scientist) CMSA-Harvard, and Google Research. mikehartleyfreedman@outlook.com}

\begin{abstract}
    Regarding the representation theorem of Kolmogorov and Arnold (KA) as an algorithm for representing or <<expressing>> functions, we test its robustness by analyzing its stability to withstand re-parameterizations of the hidden space.
    One may think of such re-parameterizations as the work of an adversary attempting to foil the construction of the KA outer function.
    We find KA to be stable under countable collections of continuous re-parameterizations, but unearth a question about the equi-continuity of the outer functions that, so far, obstructs taking limits and defeating continuous groups of re-parameterizations.
    This question on the regularity of the outer functions is relevant to the debate over the applicability of KA to the general theory of NNs.
\end{abstract}


\maketitle
\thispagestyle{empty}

\section{Introduction}

Hilbert's 13th problem \cite{wiki-Hilb-13-problem} concerns the structure of expressions for the local
motion of a root as the coefficients of a polynomial vary.
His starting point was the quadratic formula known to all school children: \(x=\frac{-b\pm\sqrt{b^2-4ac}}{2a}\) and its generalization to equations of degree $3$ and $4$.
In these formulas $x$, a function of several variables, is built as a composition of functions in which the only functions with multiple inputs are $+$ and $\cdot$. In fact, exploiting the single variable functions $\log$ and $\exp$, $\cdot$ becomes redundant.
So Hilbert asked, in modern language, about the expressivity of what would later be called a Neural Network (NN). He proposed that mathematicians devise vast generalizations of Abel's theorem on quintics, and demonstrate the limitations of compositions where the only multivariate function is $+$.
Perhaps inadvisably\footnote{If we may be so bold as to critique Hilbert, his `error' was underestimating the ability of continuous functions to confound dimensional arguments, a prime example being the space filling map \(h\colon [\,0,1\,]\to[\,0,1\,]^2\) defined by \(h(0.x_1x_2\ldots) := (0.x_1x_3\ldots, 0.x_2x_4\ldots)\), i.e. the coordinate splitting map.},
Hilbert permitted merely continuous functions within the allowed expressions and thus opened the door for Kolmogorov and Arnold (KA) to refute (a part of) Hilbert's intuition with their famous twin papers \cite{Arnold57e, Kolmogorov56}.
It took 30 years \cite{HechtNielsen1987KolmogorovsMN, 30years, Rumelhart1986LearningRB} for the computer science community to realize KA had already addressed their most basic question: expressivity.

Much debate ensued \cite{GP89, Kurkova91} over the practical implications of KA for the central questions of trainability and generalizability.
While the early consensus seemed to be that KA was not of practical value in the evolving technology, later authors \cite{KAN, Freedman24} have had a more optimistic view, focusing on the philosophy of the KA proof which hints at an optimal division of labor between the shallow and deep layers on a NN: data preparation / dynamics applied to the data.
KA gives a `universal' presentation of the data on a single hidden layer and then builds a dynamic to converge to the final layer functions ($g$) needed to represent a desired multivariate ($f$).

In this paper, we investigate the sensitivity and stability of KA NNs under continuous re-parameterizations of the hidden layer.
More precisely, we are allowed to adjust the outer function $g$, but we are not allowed to reach back and invert the re-parameterization effects on the hidden layer.

A metaphor for stability that we found useful, was to think of the intervening homeomorphism as the action of an adversary which can be overcome by a modification of the outer map(s).
KA is essentially an algorithm for building a special type (shallow, one hidden layer; but highly non-linear) of NN to represent a given continuous function. In computer science, coding theory, and the theory of error correction, the ability of an algorithm to withstand adversarial noise serves as a fundamental measure of its robustness and power. This is typically the most demanding test to which algorithms are subjected. In this spirit, we investigate KA to see how much `untapped' expressivity still resides within the basic proof method. In our context, the adversary acts on the hidden layer and its effects must be countered by a new choice of $g$.
Unlike error correcting codes, we presume knowledge of exactly how the adversary acted.

We note that in a different context the theory of NNs has already been much influenced by efforts to find and tame adversarial examples \cite{Goodfellow-Shlens-Szegedy}.
For example, NNs which easily distinguish real world examples of cats from dogs but can often be spoofed by carefully engineered pictures that look to humans exactly like `cat' or `dog' but that the NN will misclassify.
Our adversary is somewhat different; it is not acting on inputs but rather corrupting the inner workings of the NN.

Our conclusion is a bit surprising.
We find in rather general circumstances that KA persists under countable collections of re-parameterization homeomorphisms, but becomes obstructed for continuous collections. The main obstruction is the failure of equi-continuity of our final functions $g$ over the collection of re-parameterizations.
On this obstruction we are unable to answer.
A positive answer means KA can defeat re-parameterizations drawn from many continuous groups.
So, KA's exact level of stability is framed as our final Questions~\ref{q:gen}, \ref{q:out}.

\section{Background and statements}

Denote \(I := [\,0,1\,]\).

We consider only continuous functions with sup-norms, so we will omit it from the statements and definitions.
Sometimes we will use the notation \(C(X,Y)\) for such spaces of functions \(X\to Y\).
For a function \(f\colon X\to Y \subset \R^d\) we denote by \(f_i\colon X\to \R\) the composition of $f$ and the projection on the $i$-th coordinate.

The functions \(I \to I^{2n+1}\) we will call \emph{tuples}, since they may be considered as the tuples of $2n+1$ functions.
For such functions we consider the maximum norm.

For fixed \(\gamma_1, \ldots, \gamma_n \in \R\), a tuple \(\phi\colon I\to I^{2n+1}\), a function \(g\colon\R\to\R\), and a homeomorphism \(h\colon\R^{2n+1}\to\R^{2n+1}\) define the approximator \(\appx(g,h)\colon I^n\to\R\) by
\[
    \appx(g,h)(x_1,\ldots,x_n) := \sum_{j=1}^{2n+1} g\Bigl(\gamma_1h_j\bigl(\phi(x_1)\bigr) +\ldots+ \gamma_nh_j\bigl(\phi(x_n)\bigr)\Bigr).
\]

\begin{figure}[ht]
    \centering
    \begin{tikzpicture}[
    arrow/.style={
            color=black,
            draw=black,
            -latex,
                font=\fontsize{8}{8}\selectfont},
        trafo/.style={midway,font=\tiny}
    ]

    \def \dx {1}
    \def \dy {0.5}
    
    \node (I1) at (0,0) {$I$};
    \node (dots) [below=\dy of I1] {$\vdots$};
    \node (In) [below=\dy of dots] {$I$};

    \node (R1) [right=4*\dx of I1] {$\R^{2n+1}$};
    \node (ddots) [below=\dy of R1] {$\vdots$};
    \node (Rn) [below=\dy of ddots] {$\R^{2n+1}$};

    \node (RR1) [right=\dx of R1] {$\R^{2n+1}$};
    \node (dddots) [below=\dy of RR1] {$\vdots$};
    \node (RRn) [below=\dy of dddots] {$\R^{2n+1}$};

    \node (allRn) [right=\dx of dddots] {$\R^{2n+1}$};
    \node (sigma) [left=0.1*\dx of allRn] {{\scriptsize $\sum$}};
    \node (R) [right=\dx of allRn] {$\R$};
    
    \draw[->] (I1) -- (R1) node[trafo,above] {un-summed 1\textsuperscript{st} layer maps};
    \draw[->] (In) -- (Rn) node[trafo,above] {un-summed 1\textsuperscript{st} layer maps};

    \draw[->] (R1) -- (RR1) node[trafo,above] {$h$};
    \draw[->] (Rn) -- (RRn) node[trafo,above] {$h$};

    \draw[->] (RR1) to [bend right=10] (allRn.170);
    \draw[->] (RRn) to [bend left=10] (allRn.190);

    \draw[->] (allRn) -- (R) node[trafo,above] {$\sum g$};

    \draw [decorate, 
    decoration = {brace,mirror, raise=5, amplitude = 4}] (0,0.1) --  (0,-5*\dy-0.1) node[trafo, left=0.3] {$n$};

\end{tikzpicture}
    \caption{KA-style network with a homeomorphism $h$ acting on each input neuron in parallel}
    \label{fig:network-in}
\end{figure}
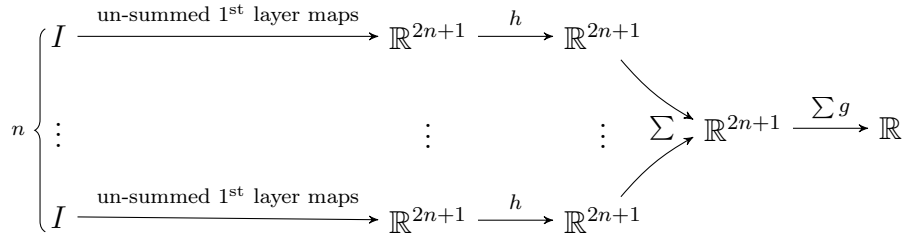

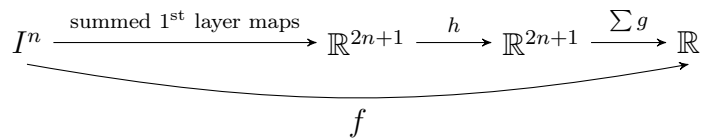
\begin{figure}[ht]
    \centering
    \begin{tikzpicture}[
    arrow/.style={
            color=black,
            draw=black,
            -latex,
                font=\fontsize{8}{8}\selectfont},
    trafo/.style={midway,font=\tiny}
    ]
    \node (In) {$I^n$};
    \node (Rnst) [right=3.5 of In] {$\R^{2n+1}$};
    \node (Rnnd) [right= of Rnst] {$\R^{2n+1}$};
    \node (R) [right= of Rnnd] {$\R$};

    \draw[->] (In) -- (Rnst) node[trafo,above] {summed 1\textsuperscript{st} layer maps};
    \draw[->] (Rnst) -- (Rnnd) node[trafo,above] {$h$};
    \draw[->] (Rnnd) -- (R) node[trafo,above] {$\sum g$};

    \draw[->] (In.south) [out=south, in=south] to [bend right=10] node[below]{$f$} (R.south);
\end{tikzpicture}
    \caption{KA-style network with a homeomorphism $h$ acting on the sum of all neuron inputs simultaneously}
    \label{fig:network-out}
\end{figure}

\begin{theorem}[proved in \S\ref{s:proof-1} below]\label{t:ka}
    Let $n>1$ be an integer.
    Let \(\mathcal{H}\) be a countable set of homeomorphisms \(\R^{2n+1}\to\R^{2n+1}\).
    For any rationally independent \(\gamma_1,\ldots,\gamma_n\in\R\) there exists a tuple \(\phi\colon I\to I^{2n+1}\) such that for any \(f\colon I^n\to\R\) and any homeomorphism \(h\in\mathcal{H}\)
    there exists a uniformly continuous function \(g\colon\R\to\R\) such that \(f = \appx(g, h)\).
\end{theorem}

In other words, Theorem~\ref{t:ka} states that we may overcome a re-parameterization (or an adversary) acting as in Figure~\ref{fig:network-in}.
Note that we cannot overcome a re-parameterization acting as in Figure~\ref{fig:network-out}, since such general homeomorphism could undo \emph{all} the preparations of the first, `blocky' layer map and render the composition \(\R^n\to\R^{2n+1}\to\R^{2n+1}\) merely an inclusion into the first $n$ coordinates of $\R^{2n+1}$.
For a specific example, one may take \(n=2\) and \(f(x,y)=xy\). It is easy to check that $f$ cannot be reconstructed at the four points \((\pm1, \pm1)\) even as \(f(x,y) = g_1(x) + g_2(y)\).
The difference between cases is that in the first case the hidden neurons receive information, it is `scrambled', and then added, while in the second case the information is added and then scrambled.

Theorem~\ref{t:ka} covers several interesting countable groups of re-parameterizations. For example, \(\mathcal{H}\) may consist of affine transformations defined by matrices and vectors with rational coefficients, or the affine group over number field, $AGL_n(\mathbb{K})$,
or the group generated by homeomorphisms whose coordinate functions are polynomials over some number field.

\begin{figure}
    \centering
    \begin{tikzpicture}[
    arrow/.style={
            color=black,
            draw=black,
            -latex,
                font=\fontsize{8}{8}\selectfont}
    ]
    \node (Phi) {$\phi_i(x)$};
    \node (Ext) [above right= of Phi] {$\phi(x) + \sqrt{2}\phi(y)$};
    \node (APhi) [below right= of Phi] {$A^i_j(\phi_i(x) + a_i)$};
    \node (ExtAPhi) [right= of APhi] {$A^i_j(\phi_i(x)+a_i) + \sqrt{2}A^i_j(\phi_i(y)+a_i)$};
    \node (AExt) [right= of Ext] {$A^i_j(\phi_i(x)+\sqrt{2}\phi_i(y)+a_i)$};

    \draw[arrow] (Phi.45) to node[above, sloped] {extension} (Ext.west);
    \draw[arrow] (Phi.315) to node[above] {$A$} (APhi.west);
    \draw[arrow] (APhi.east) to node[above] {extension} (ExtAPhi.west);
    \draw[arrow] (Ext.east) to node[above] {$A$} (AExt.west);
    \draw[<->, font=\fontsize{8}{8}\selectfont] (ExtAPhi.north) to node[above, sloped] {differ by $A^i_j(\sqrt{2}a_i)$} (AExt.south);  
\end{tikzpicture}
    \caption{Commuting of the `extension' from one to two variables and the affine transform $A$}
    \label{fig:commute}
\end{figure}
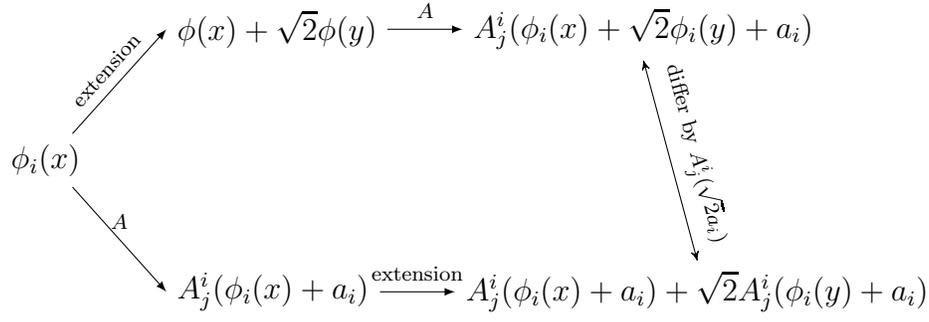

To appreciate the theorem, consider the group of invertible affine transforms \(A\colon\R^{2n+1}\to\R^{2n+1}\) defined by rational matrices \((A^i_j) \in \Q^{(2n+1)\times(2n+1)}\) and rational vectors \(a \in \Q^{2n+1}\) so that
\[
    A_j(x) = \sum_{i=1}^{2n+1} A^i_j (x_i + a_i).
\]
Then
\[
    \appx(g, A) = \sum_{j=1}^{2n+1} g\left(\sum_{i=1}^{2n+1} A^i_j\bigl(\gamma_1\phi_i(x_1) + \ldots + \gamma_n\phi_i(x_n) + (\gamma_1+\ldots+\gamma_n)a_i\bigr)\right),
\]
i.e. affine re-parameterizations commute (up to a translation) with `extension' from one to more ($n$) variables (see Figure~\ref{fig:commute} where it is shown explicitly for $n=2$).
Theorem~\ref{t:ka} implies that we may overcome an affine re-parameterization acting on all input neurons at once, as in Figure~\ref{fig:network-out}.
The latter does not hold for arbitrary re-parameterizations, as we explained right after Theorem~\ref{t:ka}.

Note that there is one more natural way to place a re-parameterization: it may act before the summation, but after the multiplication, i.e. the approximation would look like
\[
    \sum_{j=1}^{2n+1} g\Bigl(h_j\bigl(\gamma_1\phi(x_1)\bigr) +\ldots+ h_j\bigl(\gamma_n\phi(x_n)\bigr)\Bigr).
\]
Obviously, for an affine re-parameterization this is just the same function (up to a translation).
For a general case, however, we do not know how to overcome such re-parameterizations.

If there are several re-parameterizations, each acting on a different inner function, then the approximation is still possible in spite of the following result.

\begin{theorem}[proved in \S\ref{s:proof-2} below]\label{t:many-adv}
    Let $n>1$ be an integer.
    Let \(\mathcal{H}\) be a countable set of homeomorphisms \(\R^{2n+1}\to\R^{2n+1}\).
    There exist tuples \(\phi^1, \ldots, \phi^n\colon I\to I^{2n+1}\) such that for any \(f\colon I^n\to\R\) and any homeomorphisms \(h^1,\ldots,h^n\in\mathcal{H}\)
    there exists a uniformly continuous function \(g\colon\R\to\R\) such that
    \[
        f(x_1,\ldots,x_n) = \sum_{j=1}^{2n+1} g\Bigl(h^1_j\bigl(\phi^1(x_1)\bigr) + \ldots + h^n_j\bigl(\phi^n(x_n)\bigr)\Bigr)
        \qquad\text{for any}\quad x_1,\ldots,x_n \in I.
    \]
\end{theorem}

\section{Proof of Theorem~\ref{t:ka}}\label{s:proof-1}

The proof of Theorem~\ref{t:ka} we give is based on \cite{Shapiro2006topics, Lorentz-Golitschek-Makovoz}; see also \cite{Hedberg-appx}.
The idea of using a Baire category argument is due to Kahane \cite{Kahane}.

Note that in order to obtain rational independence one may take $\gamma_1=1$, \(\gamma_2=\sqrt{2}\), \(\gamma_3=\sqrt{3}\), \(\gamma_4=\sqrt{5}\), and so on, taking $\gamma_i$ to be square roots of pairwise distinct prime numbers.
Below we presume that these $\gamma_i$ are fixed (in some way).

Let \(\alpha\in(0,\frac{1}{n+1})\). 
For any \(f \colon I^n\to\R\) with \(\norm{f}=1\) and any homeomorphism \(h\colon\R^{2n+1}\to\R^{2n+1}\) define \(U_{f,h} \subset C(I, I^{2n+1})\) to be the set of tuples \(\phi\colon I\to I^{2n+1}\) such that there exists a uniformly continuous function \(g^{f,h}\colon \R\to\R\) such that \(\norm{g^{f,h}} \leq \frac{1}{n+1}\) and
\begin{equation}
    \label{eq:approx}
    \norm{f - \appx(g^{f,h}, h)} < \frac{n}{n+1} + \alpha.
\end{equation}

\begin{lemma}
\label{l:open-dense}
    Let $n>1$ be an integer and \(\alpha\in(0,\frac{1}{n+1})\).
    For any \(f \colon I^n\to \R\) with \(\norm{f}=1\) and any homeomorphism \(h\colon\R^{2n+1}\to\R^{2n+1}\) the set \(U_{f,h}\) is open dense in \(C(I,I^{2n+1})\).
\end{lemma}

\begin{proof}
    We prove the lemma for $n=2$, \(\gamma_1=1\) and \(\gamma_2=\sqrt{2}\).
    The general case differs only by the choice of constants.

    First, let us demonstrate the openness of \(U_{f,h}\).
    Suppose that \(\phi \in U_{f,h}\).
    Take \(g=g^{f,h}\) from the definition of \(U_{f,h}\).
    Set
    \[
        \varepsilon := \frac{2}{3}+\alpha - \norm{f - \appx(g, h)} > 0.
    \]
    By the uniform continuity of $g$, there is \(\delta>0\) such that \(\abs{g(x)-g(y)} < \frac{\varepsilon}{5}\) as \(\abs{x-y}<\delta\).
    Since on the cube \(I^5\) the homeomorphism $h$ is uniformly continuous, there is \(\omega>0\) such that \(\norm{h(x)-h(y)}<\frac{\delta}{1+\sqrt{2}}\) as \(\norm{x-y}<\omega\).
    Now we show that any \(\psi\) which is \(\omega\)-close to \(\phi\) also lies in \(U_{f,h}\).
    Fix such $\psi$. Then for any \(j=1,\ldots,5\) and \(x,y\in I\) we have
    \[
        \abs{h_j\bigl(\phi(x)\bigr) + \sqrt{2}h_j\bigl(\phi(y)\bigr) - h_j\bigl(\psi(x)\bigr) - \sqrt{2}h_j\bigl(\psi(y)\bigr)} < \left(1+\sqrt{2}\right)\frac{\delta}{1+\sqrt{2}} = \delta.
    \]
    For the convenience, till the end of this paragraph let us denote by \(\appx_\phi\) and \(\appx_\psi\) the approximator constructed by \(\phi\) and \(\psi\), respectively.
    Hence
    \[
        \abs{\appx_\phi(g,h)-\appx_\psi(g,h)} < 5\cdot\frac{\varepsilon}{5} = \varepsilon.
    \]
    Finally,
    \[
        \abs{f - \appx_\psi(g,h)} < \abs{f - \appx_\phi(g,h)} + \varepsilon = \frac{2}{3} + \alpha,
    \]
    which proves the openness.

    In the rest part of the proof we prove the denseness of \(U_{f,h}\).
    That is, for any \(\phi^0 \colon I\to I^5\) we must construct \(\phi \colon I\to I^5\) such that \(\norm{\phi-\phi^0}<\varepsilon\) and \eqref{eq:approx} holds for some appropriate $g^{f,h}$.
    The idea of the proof is to change the coordinate system on the one twisted by the re-parametrization~$h$, and then to repeat the classical proof.

    Let $N$ be a large positive integer to be specified later.
    \textbf{The union of red intervals of rank $i$, \(1 \leq i \leq 5\)}, is
    \[
        I \setminus \bigcup\limits_{\substack{0 \leq s \leq N \\ s \equiv i \pmod{5}}}
        \left(\frac{s}{N}, \frac{s+1}{N}\right).
    \]
    These are mostly closed intervals of length $\frac{4}{N}$ with possibly shorter intervals or points near \(\partial I = \{0,1\}\).

    Next, we specify $\phi$ somewhat indirectly by giving the following conditions on \(\phi^h := h\phi\). Since $h$ is invertible this suffices\footnote{One may think about $\phi^h$ as about original $\phi$ but in a new coordinate system, chosen by an adversary.}.

    \begin{enumerate}[label=(\alph*), ref=\alph*]
        \item\label{i:range} \( \phi^h \) has range in \( hI^5 \);

        \item\label{i:close} \(\norm{\phi^h - \phi^{0,h}} \leq \varepsilon'\), where $\varepsilon'$ is sufficiently small so that this implies \(\norm{\phi-\phi^0}<\varepsilon\);
        
        \item\label{i:rat-appx} \(\phi_j^h\) is constant and a rational number on each red interval of rank \(j=1,\ldots,5\);

        \item\label{i:rat-dif} \(\phi_j^h\) and \(\phi_{j'}^h\) assume distinct values on all red intervals, regardless of whether $j=j'$.
    \end{enumerate}
    It is obvious that~\eqref{i:range} is possible.
    Since the range \(h I^5\) is compact, we have that the inverse map \(h^{-1}\) is uniformly continuous on \(h I^5\).
    Then~\eqref{i:close} is possible.
    Now lower bound $N$ by requiring (from the uniform continuity of \(\phi\)) that \(\phi^{0,h}\) varies by at most \(\varepsilon'\) on each of the red intervals.
    Then~(\ref{i:rat-appx},\ref{i:rat-dif}) are possible since additionally \(h(0,1)^5\) is open.

    It remains to construct $g^{f,h}$.    
    A \textbf{red rectangle of rank $i$, \(i=1,\ldots,5\), in \(I^2\)}, is defined to be the Cartesian product of any two red intervals of rank $i$. Denote these rectangles by \(R_{i,1}, \ldots, R_{i,r}\) for some $r$ (probably dependent on $i$) in any order\footnote{In the case of general $n\geq2$ they would be red $n$-rectangular solids.}.

    In this paragraph we give the well-known argument in order to show that any $x \in I$ lies in red intervals of at least four different ranks; see for example \cite[Fundamental Lemma~1]{Brattka2007}.
    The red ractangles of rank $i$ are of the form
    \[
        I \cap \left[\,\frac{5r+i-4}{N}, \frac{5r+i}{N}\,\right].
    \]
    Then
    \[
        x \in I \cap \left[\,\frac{5r+i-4}{N}, \frac{5r+i}{N}\,\right] \Longleftrightarrow
        Nx \in \left[\,5r+i-4, 5r+i\,\right].
    \]
    Taking $r$ such that \(Nx \in [\,5r-4, 5r+1)\), we obtain that at least for four different $i$ the inclusion \(Nx \in \left[\,5r+i-4, 5r+i\,\right]\) holds.
    
    Hence, any \((x,y)\in I^2\) lies in red rectangles of at least three different ranks\footnote{In the case \(n\geq 2\) any point from $I^n$ lies in red $n$-rectangular solids of at least \(n+1\) different ranks.}.
    On such red rectangles \(R_{j,r}\)
    \[
        \Phi_j^h(x,y) := \phi_j^h(x) + \sqrt{2}\phi_j^h(y)
    \]
    will assume a constant value which we simply denote as \(\Phi_{j,r}^h\).

    We must further lower bound $N$ by requiring (from the uniform continuity of the function $f$) that
    \(\abs{f(x,y) - f(x',y')} < \alpha\)
    for any pairs $(x,y)$ and $(x',y')$ located in the same $R_{j,r}$.

    For each red rectangle \(R_{j,r}\) choose some point \(\ell_{j,r} \in R_{j,r}\).

    Finally, define a piecewise linear \(g^{f,h}\colon\R\to\R\) by the `forced' values \(g^{f,h}(\Phi^h_{j,r}) := \frac{1}{3}f(\ell_{j,r})\) for each red rectangle \(R_{j,r}\).
    This definition makes sense since all $\Phi_{j,r}^h$ are distinct.
    Indeed, on two different rectangles \(R_{j,r}\) and \(R_{j', r'}\) we have
    \[
        \Phi_{j,r}^h = p + \sqrt{2}q
        \quad\text{and}\quad
        \Phi_{j',r'}^h = p' + \sqrt{2}q',
    \]
    where \(p,p',q,q' \in \Q\) and at least one of the inequalities \(p\neq p'\), \(q\neq q'\) holds, and that is why \(\Phi_{j,r}^h \neq \Phi_{j',r'}^h\).
    
    Clearly, it is possible to extend \(g^{f,h}\) from its `forced' values to ensure that \(\norm{g^{f,h}} \leq \frac{1}{3}\).
    For that, one just needs to extend \(g^{f,h}\) piecewise linearly between the leftmost and the rightmost points \(\Phi^h_{j,r}\), and, on the two remaining rays, make it remain constant.
    
    Let us verify~\eqref{eq:approx}.
    Since (for any continuous \( g\colon\R\to\R \))
    \[
        \appx(g, h) =
        \sum_{j=1}^5 g \circ \Phi_j^h,
    \]
    we need to prove that
    \[
        \abs{f(x,y) - \sum_{j=1}^5 g^{f,h} \bigl( \Phi_j^h(x,y)\bigr)} < \alpha + \frac{2}{3}
        \quad\text{for any}\quad (x,y)\in I^2.
    \]
    Take any $x,y\in I$.
    As we already know, there are at least three red rectangles \(R_{j_s, r_s}\), \(s\in\{1,2,3\}\), containing the point \((x,y)\).
    For them we have \(g^{f,h}(\Phi_{j_s,r_s}^h) = \frac{1}{3}f(\ell_{j_s,r_s})\).
    These values differ from \(\frac{1}{3}f(x,y)\) by less than \(\frac{\alpha}{3}\).
    Hence
    \[
        \abs{\sum_{s=1}^3 g^{f,h}\left(\Phi^h_{j_s}(x,y)\right) - f(x,y)} < \alpha.
    \]
    The other two values \(g^{f,h}\left(\Phi^h_j(x,y)\right)\) do not exceed \(\frac{1}{3}\) by the absolute value.
\end{proof}


\begin{lemma}
\label{l:approx}
    Let $n>1$ be an integer and \(\lambda \in (\frac{2n+1}{2n+2},1)\).
    Let \(\mathcal{H} = \{h_q\}\) be a countable set of homeomorphisms \(\R^{2n+1}\to\R^{2n+1}\).
    There exists a fixed \(\phi\colon I\to I^{2n+1}\) such that given any \(f \colon I^n\to\R\) and any \(h\in\mathcal{H}\), there is a uniformly continuous \(g\colon\R\to\R\) with \(\norm{g} \leq \frac{1}{n+1}\norm{f}\) satisfying
    \[
        \norm{f - \appx(g, h)} < \lambda\norm{f}.
    \]
\end{lemma}

\begin{proof}
    W.l.o.g. assume $\norm{f}=1$ and let \(f_1, f_2, \ldots\) be an infinite sequence of functions in $C(I^n,\R)$ all of norm $1$ and dense in the unit sphere of $C(I^n,\R)$.
    Take \(\alpha := \lambda-\frac{2n+1}{2n+2}\).
    The space \(C(I, I^{2n+1})\) is a separable complete metric space, so by Lemma~\ref{l:open-dense} and the Baire category theorem,
    \[
        \bigcap_{1 \leq p,q< \infty} U_{f_p, h_q} \neq\emptyset.
    \]
    (In fact, this intersection is a dense set of `2nd category'.)
    Choose \(\phi\) from this intersection.
    For this $\phi$ choose maps $g^{f_p, h_q}$ from the definition of $U_{f_p, h_q}$.

    Now fix any $f \colon I^n\to\R$ and any \(h=h_q\). We need to choose the appropriate~$g$.
    For this, choose $f_p$ so that \(\norm{f - f_p} < \frac{1}{2n+2}\).
    
    Now \(g := g^{f_p, h_q}\) satisfies the requirements of the lemma, since
    \[
        \norm{f - \appx(g, h)} \leq
        \norm{f - f_p} + \norm{f_p - \appx(g, h)}
        <
        \frac{1}{2n+2} + \alpha + \frac{n}{n+1} = \lambda.
    \]
\end{proof}

\begin{proof}[Proof of Theorem~\ref{t:ka}]
    Take any \(\lambda \in (\frac{2n+1}{2n+2},1)\).
    Take $\phi$ from Lemma~\ref{l:approx}.
    Fix any \(f\colon I^n\to\R\) and \(h\in\mathcal{H}\).
    Lemma~\ref{l:approx} fuels a non-linear recursion (with $\phi$ fixed) in which at each step the sup-norm difference between $f$ and its \((g,h)\)-approximation is decreased by a factor of $\lambda$.
    
    Formally, let \(g_0 := 0\) and let \(g_{m+1}\) be obtained by applying Lemma~\ref{l:approx} for fixed $\phi$ to \(f-\sum\limits_{k=0}^m \appx(g_k, h)\) and $h$.
    Hence for any integer $m > 0$
    \[
        \norm{f - \sum\limits_{k=0}^m \appx(g_k, h)} <
        \lambda \norm{f - \sum\limits_{k=0}^{m-1} \appx(g_k, h)} < \ldots < \lambda^m \norm{f}.
    \]
    Since \(\norm{g_m} \leq \frac{1}{n+1} \norm{f - \sum\limits_{k=0}^{m-1} \appx(g_k, h)} \leq \frac{1}{n+1}\lambda^{m-1}\norm{f}\), the functional series \(\sum\limits_{m=0}^\infty g_m\) converges uniformly on $\R$ to a uniformly continuous function \(g\colon\R\to\R\). Then for any $j=1,\ldots,2n+1$
    \[
        \sum_{m=0}^\infty g_m\left(\gamma_1t_{j,1} + \ldots + \gamma_nt_{j,n}\right) = g\left(\gamma_1t_{j,1} + \ldots + \gamma_nt_{j,n}\right),
    \]
    where the convergence is uniform for any \(t_{j,i} \in \R\).
    Finally,
    \[
        \sum_{m=0}^\infty \appx(g_m, h) =
        \appx\left(\sum_{m=0}^\infty g_m, h\right) = \appx(g, h).
    \]
    Hence \(\norm{f - \sum\limits_{m=0}^\infty \appx(g_m, h)} = 0\), and so \(f = \appx(g, h)\).
\end{proof}

\section{Proof of Theorem~\ref{t:many-adv}}\label{s:proof-2}

In this section we give the proof of Theorem~\ref{t:many-adv}.
Since the proof is analogous to the proof of Theorem~\ref{t:ka}, the reader may skip it and go right to \S\ref{s:relate}.

For fixed tuples \(\phi^1, \ldots, \phi^n\colon I\to I^{2n+1}\), a function \(g\colon\R\to\R\), and homeomorphisms \(h^1,\ldots,h^n\colon\R^{2n+1}\to\R^{2n+1}\) define the approximator \(\appx(g,\{h^i\})\colon I^n\to\R\) by
\[
    \appx(g,\{h^i\})(x_1,\ldots,x_n) := \sum_{j=1}^{2n+1} g\Bigl(h^1_j\bigl(\phi^1(x_1)\bigr) +\ldots+ h^n_j\bigl(\phi^n(x_n)\bigr)\Bigr).
\]

Let \(\alpha\in(0,\frac{1}{n+1})\). 
For any \(f \colon I^n\to\R\) with \(\norm{f}=1\) and any homeomorphisms \(h^1,\ldots,h^n\colon\R^{2n+1}\to\R^{2n+1}\) define \(U_{f,\{h^i\}} \subset C(I, I^{2n+1})^n\) to be the set of tuples \((\phi^1\colon I\to I^{2n+1},\ldots,\phi^n\colon I\to I^{2n+1})\) such that there exists a uniformly continuous function \(g^{f,\{h^i\}}\colon \R\to\R\) such that \(\norm{g^{f,\{h^i\}}} \leq \frac{1}{n+1}\) and
\begin{equation}\label{eq:many-approx}
    \abs{f(x_1,\ldots,x_n) - \appx(g^{f, \{h^i\}}, \{h^i\})} < \frac{n}{n+1} + \alpha
\end{equation}
for any \(x_1,\ldots,x_n \in I\).

\begin{lemma}\label{l:open-dense-many}
    Let $n>1$ be an integer and \(\alpha\in(0,\frac{1}{n+1})\).
    For any \(f \colon I^n\to \R\) with \(\norm{f}=1\) and any homeomorphisms \(h^1,\ldots,h^n\colon\R^{2n+1}\to\R^{2n+1}\) the set \(U_{f,\{h^i\}}\) is open dense in \(C(I,I^{2n+1})^n\).
\end{lemma}

\begin{proof}
    First, let us demonstrate the openness of \(U_{f,h}\).
    Suppose that \(\phi \in U_{f,\{h^i\}}\).
    Take \(g=g^{f,\{h^i\}}\) from the definition of \(U_{f,\{h^i\}}\).
    Set
    \[
        \varepsilon := \frac{n}{n+1}+\alpha - \norm{f - \appx(g, \{h^i\})} > 0.
    \]
    By the uniform continuity of $g$, there exists \(\delta>0\) such that \(\abs{g(x)-g(y)} < \frac{\varepsilon}{2n+1}\) as \(\abs{x-y}<\delta\).
    Since on the cube \(I^{2n+1}\) the homeomorphisms $h^i$ are uniformly continuous, there is common \(\omega>0\) such that \(\norm{h^i(x)-h^i(y)}<\frac{\delta}{n}\) as \(\norm{x-y}<\omega\).
    Now we show that any \(\psi\) which is \(\omega\)-close to \(\phi\) also lies in \(U_{f,\{h^i\}}\).
    Fix such $\psi$. Then for any \(j=1,\ldots,2n+1\) and \(x_1,\ldots,x_n\in I\) we have
    \[
        \abs{h^1_j\bigl(\phi^1(x_1)\bigr) +\ldots+ h^n_j\bigl(\phi^n(x_n)\bigr) - h^1_j\bigl(\psi^1(x_1)\bigr) -\ldots- h^n_j\bigl(\psi^n(x_n)\bigr)} < n\frac{\delta}{n} = \delta.
    \]
    For the convenience, till the end of this paragraph let us denote by \(\appx_\phi\) and \(\appx_\psi\) the approximator constructed by \(\phi\) and \(\psi\), respectively.
    Hence
    \[
        \abs{\appx_\phi(g,h)-\appx_\psi(g,h)} < (2n+1)\frac{\varepsilon}{2n+1} = \varepsilon.
    \]
    Finally,
    \[
        \abs{f - \appx_\psi(g,h)} < \abs{f - \appx_\phi(g,h)} + \varepsilon = \frac{n}{n+1} + \alpha,
    \]
    which proves the openness.
    
    In the rest part of the proof we prove the denseness.
    That is, for any \(\phi^{0,1},\ldots,\phi^{0,n} \colon I\to I^{2n+1}\) we must construct \(\phi^1,\ldots,\phi^n \colon I\to I^{2n+1}\) such that \(\norm{\phi^i-\phi^{0,i}}<\varepsilon\) for each $i$ and \eqref{eq:many-approx} holds for some appropriate $g^{f,\{h^i\}}$.
    
    The proof differs from the proof of Lemma~\ref{l:open-dense} only in the construction of functions \(\phi^i\).
    Take red intervals of rank $i \equiv 0, \ldots, 2n \pmod{2n+1}$ from the proof of Lemma~\ref{l:open-dense}.
    Take \(\gamma_i\) to be square roots of pairwise distinct prime numbers.
    Now we specify $\phi^i$ somewhat indirectly by giving the following conditions on \(\phi^{h^i} := h^i\phi^i\). Since $h^i$ are invertible this suffices.

    \begin{enumerate}[label=(\alph*'), ref=\alph*']
        \item\label{i:m:range} \( \phi^{h^i} \) has range in \( h^iI^{2n+1} \) for each $i=1,\ldots,n$;

        \item\label{i:m:close} \(\norm{\phi^{h^i} - \phi^{0,h^i}} \leq \varepsilon'\), where $\varepsilon'$ is sufficiently small so that this implies \(\norm{\phi^i-\phi^{0,i}}<\varepsilon\);
        
        \item\label{i:m:rat-appx} \(\phi_j^{h^i}\) is constant and a number of the kind \(q\gamma_i\) on each red interval of rank \(j=1,\ldots,2n+1\), where \(q\in\Q\setminus \{0\}\);

        \item\label{i:m:rat-dif} \(\phi_j^{h^i}\) and \(\phi_{j'}^{h^i}\) assume distinct values on all red intervals, regardless of whether $j=j'$.
    \end{enumerate}
    Here we need to lower bound $N$ by requiring (from the uniform continuity of \(\phi^i\)) that \(\phi^{0,h^i}\) varies by at most \(\varepsilon'\) on each of the red intervals for any \(i=1,\ldots,n\).

    Now taking red $n$-rectangular solids of rank $i$ as the Cartesian products of red intervals of rank $i$, we observe that any point \((x_1,\ldots,x_n)\in I^n\) lies in red $n$-rectangular solids of at least \(n+1\) different ranks.
    On such $n$-rectangular solids \(R_{j,r}\)
    \[
        \Phi^{\{h^i\}}(x_1,\ldots,x_n) := \phi^{h^1}(x_1) + \ldots + \phi^{h^n}(x_n)
    \]
    assume a constant value which we simply denote as \(\Phi^{\{h^i\}}_{j,r}\).
    
    We must further lower bound $N$ by requiring (from the uniform continuity of the function $f$) that
    \(\abs{f(x) - f(x')} < \alpha\)
    for any tuples $x,x'\in I^n$ located in the same $R_{j,r}$.

    For each red $n$-rectangular solids \(R_{j,r}\) choose some point \(\ell_{j,r} \in R_{j,r}\).

    Finally, define a piecewise linear \(g^{f,\{h^i\}}\colon\R\to\R\) by the `forced' values \(g^{f,\{h^i\}}(\Phi^{\{h^i\}}_{j,r}) := \frac{1}{n+1}f(\ell_{j,r})\) for each red $n$-rectangular solids \(R_{j,r}\).
    This definition makes sense since all $\Phi_{j,r}^{\{h^i\}}$ are distinct.
    Indeed, on two different red $n$-rectangular solids \(R_{j,r}\) and \(R_{j', r'}\) we have
    \[
        \Phi_{j,r}^{\{h^i\}} = \sum_{i=1}^n q_i\gamma_i
        \quad\text{and}\quad
        \Phi_{j',r'}^h = \sum_{i=1}^n q'_i\gamma_i,
    \]
    where \(q_i,q'_i \in \Q\) and at least one of the inequalities \(q_i\neq q_i'\) holds, and that is why \(\Phi_{j,r}^{\{h^i\}} \neq \Phi_{j',r'}^{\{h^i\}}\).
    
    Again, extend \(g^{f,h}\) from its `forced' values to ensure that \(\norm{g^{f,h}} \leq \frac{1}{n+1}\).
    
    Let us verify~\eqref{eq:many-approx}.
    Since (for any continuous \( g\colon\R\to\R \))
    \[
        \appx(g, \{h^i\}) =
        \sum_{j=1}^{2n+1} g \circ \Phi_j^{\{h^i\}},
    \]
    we need to prove that
    \[
        \abs{f(x) - \sum_{j=1}^{2n+1} g^{f, \{h^i\}} \bigl( \Phi_j^{\{h^i\}}(x)\bigr)} < \alpha + \frac{n}{n+1}
        \quad\text{for any}\quad x\in I^n.
    \]
    Take any $x\in I^n$.
    As we already know, there exist at least $n+1$ red $n$-rectangular solids \(R_{j_s, r_s}\), \(s\in\{1,\ldots,n+1\}\), containing the point \(x\).
    For them we have \(g^{f,\{h^i\}}(\Phi_{j_s,r_s}^{\{h^i\}}) = \frac{1}{n+1}f(\ell_{j_s,r_s})\).
    These values differ from \(\frac{1}{n+1}f(x)\) by less than \(\frac{\alpha}{n+1}\).
    Hence
    \[
        \abs{\sum_{s=1}^{n+1} g^{f,\{h^i\}}\left(\Phi^{\{h^i\}}_{j_s}(x)\right) - f(x)} < \alpha.
    \]
    The other $n$ values \(g^{f,\{h^i\}}\left(\Phi^{\{h^i\}}_j(x)\right)\) do not exceed \(\frac{1}{n+1}\) by the absolute value.  
\end{proof}

\begin{lemma}
\label{l:approx-many}
    Let $n>1$ be an integer and \(\lambda \in (\frac{2n+1}{2n+2},1)\).
    Let \(\mathcal{H} = \{h_q\}\) be a countable set of homeomorphisms \(\R^{2n+1}\to\R^{2n+1}\).
    There exist fixed \(\phi^1,\ldots,\phi^n\colon I\to I^{2n+1}\) such that given any \(f \colon I^n\to\R\) and any \(h^1,\ldots,h^n\in\mathcal{H}\), there is a uniformly continuous \(g\colon\R\to\R\) with \(\norm{g} \leq \frac{1}{n+1}\norm{f}\) satisfying
    \[
        \norm{f - \appx(g, \{h^i\})} < \lambda\norm{f}.
    \]
\end{lemma}

\begin{proof}
    W.l.o.g. assume $\norm{f}=1$ and let \(f_1, f_2, \ldots\) be an infinite sequence of functions in $C(I^n,\R)$ all of norm $1$ and dense in the unit sphere of $C(I^n,\R)$.
    Take \(\alpha := \lambda-\frac{2n+1}{2n+2}\).
    The space \(C(I, I^{2n+1})^n\) is a separable complete metric space, so by Lemma~\ref{l:open-dense-many} and the Baire category theorem,
    \[
        \bigcap_{1 \leq p,q< \infty} U_{f_p, \{h^i\}_q} \neq\emptyset.
    \]
    (In fact, this intersection is a dense set of `2nd category'.)
    Choose \(\phi = (\phi^1,\ldots,\phi^n)\) from this intersection.
    For this $\phi$ choose maps $g^{f_p, \{h^i\}_q}$ from the definition of $U_{f_p, \{h^i\}_q}$.

    Now fix any $f \colon I^n\to\R$ and any \(\{h^i\}=\{h^i\}_q\). We need to choose the appropriate~$g$.
    For this, choose $f_p$ so that \(\norm{f - f_p} < \frac{1}{2n+2}\).
    
    Now \(g := g^{f_p, \{h^i\}_q}\) satisfies the requirements of the lemma, since
    \[
        \norm{f - \appx(g, \{h^i\})} \leq
        \norm{f - f_p} + \norm{f_p - \appx(g, \{h^i\})}
        <
        \frac{1}{2n+2} + \alpha + \frac{n}{n+1} = \lambda.
    \]
\end{proof}

\begin{proof}[Proof of Theorem~\ref{t:many-adv}]
    Take any \(\lambda \in (\frac{2n+1}{2n+2},1)\).
    Take $\phi$ from Lemma~\ref{l:approx-many}.
    Fix any \(f\colon I^n\to\R\) and \(h^1,\ldots,h^n\in\mathcal{H}\).
    
    Let \(g_0 := 0\) and let \(g_{m+1}\) be obtained by applying Lemma~\ref{l:approx-many} for fixed $\phi$ to \(f-\sum\limits_{k=0}^m \appx(g_k, \{h^i\})\) and $\{h^i\}$.
    Hence for any integer $m > 0$
    \[
        \norm{f - \sum\limits_{k=0}^m \appx(g_k, \{h^i\})} <
        \lambda \norm{f - \sum\limits_{k=0}^{m-1} \appx(g_k, \{h^i\})} < \ldots < \lambda^m \norm{f}.
    \]
    Since \(\norm{g_m} \leq \frac{1}{n+1} \norm{f - \sum\limits_{k=0}^{m-1} \appx(g_k, \{h^i\})} \leq \frac{1}{n+1}\lambda^{m-1}\norm{f}\), the functional series \(\sum\limits_{m=0}^\infty g_m\) converges uniformly on $\R$ to a uniformly continuous function \(g\colon\R\to\R\).
    Then
    \[
        \sum_{m=0}^\infty \appx(g_m, \{h^i\}) =
        \appx\left(\sum_{m=0}^\infty g_m, \{h^i\}\right) = \appx(g, \{h^i\}).
    \]
    Hence \(\norm{f - \sum\limits_{m=0}^\infty \appx(g_m, \{h^i\})} = 0\), and so \(f = \appx(g, \{h^i\})\).
\end{proof}

\section{Related work and positioning}\label{s:relate}

See the overview \cite{Vitushkin-overview} for the history of KA, its discussion and applications.
The KA was well studied in the spirit of strengthening the restrictions imposed on the functions:
\begin{itemize}[nosep]
    \item in \cite{Vitushkin-1954}, it was shown that KA fails if all the functions are required to be continuously differentiable,

    \item in \cite{Fridman}, it was shown that all the inner layer functions \(\phi\) could be made $1$-Lipschitz continuous,

    \item it is known \cite{Lorentz-Golitschek-Makovoz, Hedberg-appx} that all the inner layer functions \(\phi\) could be made strictly increasing.
\end{itemize}

Note that in \cite{Braun-Griebel-2009} the first constructive proof of KA theorem was given.

Instability of KA representations was known qualitatively.
Our paper is probably the first to investigate the uniform stability of KA under the effects of collections of re-parameterizations, acting on the hidden layer.

We leave the reader with the following open questions.

\begin{question}\label{q:gen}
    Do the analogues of Theorems~\ref{t:ka}, \ref{t:many-adv} hold when the collection of re-parameterizations is the entire homeomorphism group of \(\R^{2n+1}\)?
\end{question}

\begin{question}\label{q:out}
    Do the analogues of Theorems~\ref{t:ka}, \ref{t:many-adv} hold for re-parameterizations acting \textbf{after} the summation, as diagramed in Figure~\ref{fig:network-out}, for any interesting class of re-parameterizations beyond the cases of: 1. Countable subsets of the affine group, and 2. The continuous group of translations \(\{T\}\) as discussed in the final paragraph of \S\ref{s:concl}.Conclusions (below)?
    Both countable and continuous families of \(C^2\)-diffeomorphisms look like an interesting case to consider.
\end{question}

\section{Conclusions}\label{s:concl}

All proofs of KA using the Baire category theorem, including ours, easily pass from a countable dense collection $\{f\}$ of functions for which a `working' $\phi$ has been found to all functions but taking a limit.
It may therefore come as a surprise that we are not able to similarly take a limit to pass from a countable collection $\mathcal{H}$ of re-parameterizations to its closure.
This would be the hoped-for conclusion. What goes wrong?
The problem is that our rule for producing the outer function $g^{f,h}$ depends delicately on the location of the `forced' values on the real line. The function $g^{f,h}$ at these values (both in the approximation step and after convergence) bounces around between the various \(\frac{1}{3}f(\ell_{j,r})\), which initially means within \([\,-\frac{1}{3}, \frac{1}{3}\,]\).
If such large transitions occur at near and nearer `forced' values the Lipshitz constant of $g^{f,h}$ will diverge.
As the re-parameterization $h$ varies (think of rotating the plane and projecting) the `forced' values may cross each other.
This means that the functions \(\{g^{f,h}\}\) are not necessarily equi-continuous w.r.t. $h$, and this turns out to be the main obstruction for the step from countable collections of homeomorphisms to continuous ones.

To give a very simple instance of this problem, consider a re-parameterization that acts on $\R^5$ merely by Euclidian translation. So, \(\R^5=\{T\}\) is (also) our group of allowed re-parameterizations.
If one translates any given coordinate axis, the `forced' values will not cross and, trivially, the re-parameterization may be defeated by simply pre-composing  \(\{g^T_1,\ldots,g^T_5\}\) by the inverse translation \(T^{-1}\)~--- \textbf{if} we are allowed five independent outer functions, one for each hidden layer neuron.
If we continue, though, to maintain the initial rules of our game, and ask for a single outer function \(g^T\), then the independent axial translations of the `forced' values, now projected onto a single Real axis, will in general cross, causing a divergence of \(\mathop{\text{Lip}}(g^T)\) and a loss of equi-continuity. This is the simplest context in which we do not know how to answer our question.
As explained in our discussion of affine re-parameterizations, translations \(\{T\}\) it makes no difference whether the re-parameterization action occurs before or after summation is taken at the hidden layer.

There is another natural question concerning the computability of all functions.
We do not give the precise definition of a computable function \(\R\to\R\) which can be found, for example, in \cite{Brattka2007, wiki-Computable-func}.
Analogously to \cite{HechtNielsen1987KolmogorovsMN, Brattka2007}, we pose the following final question.

\begin{question}
    Does the class of functions \(I^n \to \R\), implementable by neural networks with a single hidden layer and computable activation functions, computable weights, and a computable adversary acting as in Figure~\ref{fig:network-in}, coincide with the class of all computable functions \(I^n\to\R\)?
\end{question}

\printbibliography

@article{30years,
  author={Widrow, B. and Lehr, M.A.},
  journal={Proceedings of the IEEE}, 
  title={30 years of adaptive neural networks: perceptron, {M}adaline, and backpropagation}, 
  year={1990},
  volume={78},
  number={9},
  pages={1415-1442},
  doi={10.1109/5.58323}
}

@article{Arnold57e,
    author = {Arnold, Vladimir},
    title = {On functions of three variables},
    journal = {Proceedings of the USSR Academy of Sciences, 114},
    year = {1957},
    pages = {679--681},
    note = {English translation: Amer. Math. Soc. Transl., "28: Sixteen Papers on Analysis" (1963), pp. 51–-54.}
}

@article{Braun-Griebel-2009,
  title = {On a Constructive Proof of Kolmogorov’s Superposition Theorem},
  author = {J\"{u}rgen Braun and Michael Griebel},
  journal = {Constructive Approximation},
  year = {2009},
  volume = {30},
  number = {3},
  pages = {653--675},
  doi = {10.1007/s00365-009-9054-2}
}

@article{Freedman24,
    author = {Freedman, Michael H.},
    title = {The proof of {K}olmogorov-{A}rnold May Illuminate {N}eural {N}etwork Learning},
    journal = {arXiv:2410.08451 preprint},
    year = {2024}
}

@article{Fridman,
    author = {B.~L.~Fridman},
    title = {An improvement in the smoothness of the functions in A.\,N.~Kolmogorov's theorem on superpositions},
    journal = {Dokl. Akad. Nauk SSSR},
    year = {1967},
    volume = {177},
    issue = {5},
    pages = {1019-1022},
    url = {http://mi.mathnet.ru/dan33525}
}

@article{GP89,
  author={Girosi, Federico and Poggio, Tomaso},
  journal={Neural Computation}, 
  title={Representation Properties of {N}etworks: {K}olmogorov's {T}heorem Is Irrelevant}, 
  year={1989},
  volume={1},
  number={4},
  pages={465-469},
  doi={10.1162/neco.1989.1.4.465}
}

@article{Goodfellow-Shlens-Szegedy,
    author = {Ian J Goodfellow and Jonathon Shlens and Christian Szegedy},
    title = {Explaining and harnessing adversarial examples},
    journal = {arXiv:1412.6572 preprint},
    year = {2014}
}

@article{Hedberg-appx,
    author = {Hedberg, T.},
    title = {The Kolmogorov superposition theorem, Appendix II to H.S.Shapiro, Topics in Approximation Theory},
    journal = {Lecture Notes in Math.},
    year = {1971},
    volume = {187},
    pages = {267--275}
}

@article{KAN,
    author = {Liu, Ziming and Wang, Yixuan and Vaidya, Sachin and Ruehle, Fabian and Halverson, James and Solja\v{c}i\'{c}, Marin and Hou, Thomas Y. and Tegmark, Max},
    title = {{KAN}: {K}olmogorov-{A}rnold {N}etworks},
    journal = {arXiv:2404.19756 preprint},
    year= {2024}
}

@article{Kahane,
    author = {Kahane, J.P.},
    title = {Sur le theoreme de superposition de {K}olmogorov},
    journal = {J. Approximation Theory},
    year = {1975},
    volume = {13},
    pages = {229--234}
}

@article{Kolmogorov56,
    author = {Kolmogorov, Andrey},
    title = {On the representation of continuous functions of several variables by superpositions of continuous functions of a smaller number of variables (in Russian)},
    journal = {Proceedings of the USSR Academy of Sciences, 108},
    year = {1956},
    pages = {179--182},
    note = {English translation: Amer. Math. Soc. Transl., "17: Twelve Papers on Algebra and Real Functions" (1961), pp. 369–373.}
}

@article{Kurkova91,
    author = {K\r{u}rkov\'{a}, V\v{e}ra},
    title = {Kolmogorov's {T}heorem Is Relevant},
    journal = {Neural Comput.  Winter;3(4)},
    year = {1991},
    pages = {617--622},
    doi = {10.1162/neco.1991.3.4.617},
    note = {PMID: 31167327}
}

@article{Lorentz-Golitschek-Makovoz,
    author = {G.G. Lorentz and M.v. Golitschek and Y. Makovoz},
    title = {Constructive Approximation: Advanced Problems},
    journal = {Springer,  New York,},
    year = {1996}
}

@book{Shapiro2006topics,
  title={Topics in {A}pproximation {T}heory},
  author={Shapiro, H.S.},
  isbn={9783540364979},
  series={Lecture Notes in Mathematics},
  url={https://books.google.ru/books?id=3jl6CwAAQBAJ},
  year={2006},
  publisher={Springer Berlin Heidelberg}
}

@article{Rumelhart1986LearningRB,
  title={Learning representations by back-propagating errors},
  author={David E. Rumelhart and Geoffrey E. Hinton and Ronald J. Williams},
  journal={Nature},
  year={1986},
  volume={323},
  pages={533-536},
  url={https://api.semanticscholar.org/CorpusID:205001834}
}

@article{Vitushkin-1954,
    author={A.~G.~Vitushkin},
    title={On Hilbert's thirteenth problem},     
    journal = {Dokl. Akad. Nauk SSSR (Russian)},
    volume={96},
    pages={701-704},
    year = {1954}
}

@article{Vitushkin-overview,
    author={A.~G.~Vitushkin},
    title={On Hilbert's thirteenth problem and related questions},
    journal={Russian Math. Surveys},
    year={2004},
    volume={59},
    issue={1},
    pages={11--25},
    doi={10.1070/RM2004v059n01ABEH000698}
}

@Inbook{Brattka2007,
    author="Brattka, Vasco",
    editor="Charpentier, {\'E}ric
    and Lesne, Annick
    and Nikolski, Nikola{\"i} K.",
    title="From Hilbert's 13th Problem to the theory of neural networks: constructive aspects of Kolmogorov's Superposition Theorem",
    bookTitle="Kolmogorov's Heritage in Mathematics",
    year="2007",
    publisher="Springer Berlin Heidelberg",
    address="Berlin, Heidelberg",
    pages="253--280",
    isbn="978-3-540-36351-4",
    doi="10.1007/978-3-540-36351-4_13"
}

@inproceedings{HechtNielsen1987KolmogorovsMN,
  title={Kolmogorov's Mapping Neural Network Existence Theorem},
  author={Robert Hecht-Nielsen},
  year={1987},
  url={https://api.semanticscholar.org/CorpusID:118526925}
}

@misc{wiki-Hilb-13-problem,
    title={Hilbert's 13th problem},
    key={Wiki},
    url={https://en.wikipedia.org/wiki/Hilbert%27s_thirteenth_problem}
}

@misc{wiki-Computable-func,
title={Computable real function},
key={Wiki},
url={https://en.wikipedia.org/wiki/Computable_real_function}
}

\end{document}